\documentclass{amia}
\usepackage{lineno}
\usepackage{booktabs}
\usepackage{enumitem}
\usepackage{xspace}
\usepackage{tcolorbox}
\tcbset{
  promptbox/.style={
    colback=blue!2!white,
    colframe=blue!60!black,
    boxrule=0.4pt, arc=2pt,
    left=8pt, right=8pt, top=6pt, bottom=6pt,
    fonttitle=\bfseries
  }
}

\newcommand{\header}[1]{\noindent \textbf{\textit{#1}}}
\newcommand{\modelname}[0]{LLM-MINE\xspace}

\begin{document}
\title{LLM-MINE: Large Language Model based Alzheimer's Disease and Related Dementias Phenotypes Mining from Clinical Notes}

\author{Mingchen Shao, BS$^1$, Yuzhang Xie, MS$^2$, Carl Yang, PhD$^2$, Jiaying Lu, PhD$^2$\footnote{Corresponding Author: Jiaying Lu (jiaying.lu@emory.edu).} }

\institutes{
    $^1$Northwestern University, Evanston, IL; $^2$Emory University, Atlanta, GA
}

\maketitle   

\section*{Abstract}
Accurate extraction of Alzheimer's Disease and Related Dementias (ADRD) phenotypes from electronic health records (EHR) is critical for early-stage detection and disease staging. However, this information is usually embedded in unstructured textual data rather than tabular data, making it difficult to be extracted accurately. We therefore propose \modelname, a Large Language Model-based phenotype mining framework for automatic extraction of ADRD phenotypes from clinical notes. Using two expert-defined phenotype lists, we evaluate the extracted phenotypes by examining their statistical significance across cohorts and their utility for unsupervised disease staging. Chi-square analyses confirm statistically significant phenotype differences across cohorts, with memory impairment being the strongest discriminator. Few-shot prompting with the combined phenotype lists achieves the best clustering performance (ARI=0.290, NMI=0.232), substantially outperforming biomedical NER and dictionary-based baselines. Our results demonstrate that LLM-based phenotype extraction is a promising tool for discovering clinically meaningful ADRD signals from unstructured notes.
\section*{Introduction}
Alzheimer's Disease and Related Dementias (ADRD) encompasses a spectrum of progressive neurocognitive disorders that develop through identifiable stages, typically beginning with normal cognition (CN), progressing to mild cognitive impairment (MCI), and eventually advancing to overt dementia~\cite{Sperling2011PreclinicalAD}. 
ADRD affects more than 6 million people in the U.S.~\cite{alzheimer2025} and over 55 million people worldwide~\cite{WHO_Dementia}. Currently, no treatments exist to prevent or halt the progression of ADRD~\cite{NINDS_ADRD_Focus}. 
Throughout the ADRD continuum, individuals exhibit varying phenotypes of cognitive deficits, encompassing domains such as memory, executive function, language, and other cognitive abilities~\cite{edition2013diagnostic}. These phenotypes serve as clinical markers that aid in detecting early cognitive decline, monitoring disease progression, and distinguishing between different stages of impairment~\cite{Grand2011DementiaCare}. Precise characterization of dementia phenotypes is therefore essential for timely diagnosis, risk stratification, and informing appropriate clinical management strategies.

The phenotypic characterization of ADRD is inherently multidimensional~\cite{lyketsos2022standardizing}, spanning domains that include demographics, disease manifestations (e.g., comorbidities, symptoms, functional decline), clinical management (e.g., medication), physiological biomarkers (e.g., imaging, labs results, vital signs), and genomic profiles. 
Electronic health records (EHRs) offer a rich, real-world data source at scale \cite{xie2025hypkg}. However, critical ADRD phenotypes are seldom encoded in structured EHR fields, rendering them inaccessible for analytics~\cite{Walling2023DementiaEHRPhenotypes}. For example, cognitive assessments and clinician impressions are often documented in free-text notes or ad hoc tools such as handwritten MMSE or MoCA scores~\cite{Wang2025CognitiveAssessmentNLP_EHR} rather than stored in consistent structured tabular formats. Similarly, longitudinal data essential for tracking progression such as cognitive function, neuropsychiatric symptoms, and neurological signs are rarely captured systematically~\cite{Oh2023PhenotypesAD}. 
Consequently, critical phenotype data required for ADRD research remains locked within narrative clinical documentation.
Recovering these phenotypes through automated mining is critical for monitoring disease progression, and understanding underlying disease mechanism.

For these reasons, we focus on clinical notes as the primary data source for phenotype mining. Clinical notes contain detailed and nuanced descriptions of cognitive status, symptom trajectories and provider impressions that are essential for characterizing ADRD progression~\cite{Oh2023PhenotypesAD}. In addition, structured EHR data can be incomplete or unavailable in many low-resource settings due to limited financial, technological, and organizational infrastructure~\cite{Ruckdeschel2023SmokingHistoryEHR}; in such contexts, documentation may rely heavily on handwritten records or locally stored digital notes. Methods that can robustly extract phenotypes from unstructured text therefore hold broad relevance across diverse healthcare settings.
However, extracting meaningful phenotypes from clinical notes presents substantial challenges. Unlike structured data, free-text requires extensive preprocessing and relies on advanced natural language understanding to capture clinically relevant information~\cite{Huckvale2019DigitalPhenotyping}. Prior work has commonly relied on predefined vocabularies or traditional natural language processing (NLP) tools~\cite{Oh2023PhenotypesAD, Ruckdeschel2023SmokingHistoryEHR}. While effective in some settings, these approaches are often hard to customize and may not generalize well to complex or institution-specific clinical language~\cite{Bilal2025NLP_CancerEHR_Review}.

To address these challenges, we propose \modelname,
a large language model-based framework for extracting ADRD-related phenotypes from clinical notes. Unlike traditional NLP approaches that require task-specific training data and extensive feature engineering, \modelname leverages the advanced natural language understanding and domain adaptation capabilities of LLMs.
\modelname is both (1) \textit{user-customizable}: allowing researchers to adapt their own phenotype list without retraining the model; and (2) \textit{training-efficient}: requiring minimal task-specific examples rather than large annotated corpus. Specifically, we prompt an LLM to extract ADRD phenotypes predefined by user-specified list from input text.
We evaluate \modelname on a large, real-world cohort comprising 8.4K CN, 800 MCI, and 8.4K ADRD patients from the MIMIC-IV database~\cite{2023MIMIC}. Given the absence of ground truth annotations, we design a comprehensive, multi-faceted evaluation protocol to indirectly validate the extracted phenotypes. 
Our contributions are three-fold: 
(1) \textit{Statistical validation}:  we apply chi-square tests to identify significant differences in phenotype prevalence across the CN, MCI, and ADRD groups, demonstrating that \modelname captures clinically meaningful variations aligned with disease progression. 
(2) \textit{Unsupervised clustering}: we show that the extracted phenotypes cluster patients in a manner consistent with ADRD stages, providing evidence that our method recovers underlying disease structure, even without supervised training.
(3) \textit{Superior performance over baselines}: we systematically compare \modelname against dictionary-based and NER-based approaches, demonstrating significant improvements in phenotype extraction accuracy.
Our results establish \modelname as a promising tool for unlocking clinically meaningful phenotypes from unstructured narratives.

\section*{Related Work}
\header{ADRD Phenotyping Using Structured EHR Data.}
Recent studies have used phenotypes present within structured EHR data to predict the risk of developing ADRD~\cite{Li2023EarlyPredictionADRD_EHR, Tang2024EHRKnowledgeNetworksADPrediction,xie2025psci}. 
Routinely recorded EHR variables such as demographics, medical diagnoses, vital signs, and medications are commonly used as ADRD phenotypes to build risk prediction models~\cite{Walling2023DementiaEHRPhenotypes}. However, many critical ADRD phenotypes, such as cognitive assessments and clinician impressions are seldom encoded in structured EHR fields \cite{Walling2023DementiaEHRPhenotypes}. This limitation motivates our focus on unstructured clinical notes as the primary data source for phenotype mining.
Other important ADRD prognostic phenotypes that exist in the broader context of EHR include brain imaging data and genetic data. While these modalities offer valuable insights, we restrict our focus to phenotypes documented in clinical notes.



\header{NLP Models for ADRD Phenotype Extraction.}
\label{sec:related sec3}
Extracting ADRD phenotypes from unstructured clinical notes remains a key challenge in medical informatics. The earliest approaches relied on rule-based NLP pipelines. For instance, Chen et al.~\cite{chen2023assess} developed regular expressions and pattern-matching rules to identify 85 cognitive tests and 11 biomarkers from clinical text. While interpretable, rule-based methods struggle to generalize beyond predefined patterns, often missing nuanced phrasing and complex semantic relationships.
To address these limitations, subsequent work adopted deep learning-based models, particularly transformers pre-trained on biomedical corpora. Cheng et al.~\cite{cheng2025high} fine-tuned BioBERT to classify sentences from memory clinic notes across seven cognitive domains (e.g., memory, executive function, language), identifying whether they described impaired or intact symptoms.
Most recently, large language models (LLMs) have demonstrated significant advances in understanding complex clinical narratives~\cite{xie2024promptlink}. For example, Shao et al.~\cite{Shao_2025} used LLMs to extract socioeconomic phenotypes from heart failure patients' notes, finding significant associations with 30-day readmission risks. However, despite these advances, the application of LLMs to ADRD-specific phenotype extraction remains underexplored—a gap this work aims to address.

\section*{Method}
\header{Preliminaries.} We first mathematically define the task of this study.
\noindent\textit{Definition 1 (Phenotype Extraction from Clinical Notes).}
Given a patient's note $\mathbf{T}=\{t_1, \dots, t_{|\mathbf{T}|}\}$ consisting of a sequence of $|\mathbf{T}|$ tokens, the goal is to utilize a LLM $f_{\theta}$ to extract certain ADRD-related phenotypes $\textbf{P}=\{p_1, \dots, p_{|\mathbf{P}|}\}$ that is from a predefined phenotype list. 
Therefore, the mining process is defined as $\mathbf{P}=f_\theta(\mathbf{T},\mathbf{I_{P}})$, where $\mathbf{I_{P}}$ denotes the phenotype specific mining prompt for LLM.

\begin{wraptable}{r}{0.45\textwidth}
\centering
\captionof{table}{\textbf{Characteristics of constructed cohort.}}
\label{tab:adrd_characteristics}
\vspace{-5pt}
\resizebox{0.95\linewidth}{!}{%
\begin{tabular}{lccc}
\toprule
\textbf{Group} & \textbf{CN} & \textbf{MCI} & \textbf{ADRD} \\
\midrule
\textbf{\#Patients} & 8,372 & 841 & 8,327 \\
\textbf{\#D/C Notes} & 11,408 & 992 & 13,675 \\
\hline
\textbf{\textit{Age}} & & & \\
\quad Mean (SD) & 64.0 (12.9) & 71.1 (15.2) & 80.2 (9.1) \\
\textbf{\textit{Sex}} \\
\quad Female (\%) & 4,337 (51.8\%) & 433 (51.5\%) & 4,944 (59.4\%) \\
\quad Male (\%) & 4,035 (48.2\%) & 408 (48.5\%) & 3,383 (40.6\%) \\
\textbf{\textit{Ethnicity}} \\
\quad Hispanic (\%) & 431 (5.2\%) & 34 (4.0\%) & 281 (3.4\%) \\
\quad Non-Hisp (\%) & 7903 (94.4\%)& 777 (92.4\%)& 7622 (91.5\%)\\ 
\quad Unknown Ethnicity (\%) & 38 (0.5\%)& 30 (3.6\%)& 424 (5.1\%)\\
\textbf{\textit{Race}} \\
\quad  White (\%) & 6,003 (71.7\%) & 625 (74.3\%) & 6,142 (73.8\%) \\
\quad  Black (\%) & 1,432 (17.1\%) & 113 (13.4\%) & 1,036 (12.4\%) \\
\quad  Asian (\%) & 236 (2.8\%) & 17 (2.0\%) & 191 (2.3\%) \\
\quad  Others (\%) & 298 (3.6\%) & 23 (2.7\%) & 349 (4.2\%) \\
\quad  Unknown (\%) & 38 (0.5\%)& 30 (3.6\%)& 424 (5.1\%)\\
\bottomrule
\end{tabular}%
}
\end{wraptable}
\noindent
\header{Data Source and Cohort Construction.}
This study uses unstructured clinical notes from MIMIC-IV~\cite{2023MIMIC}, a large open source healthcare database that contains information for patients admitted to either emergency department or critical care units. 
To investigate the progression of neurocognitive decline, we construct a cohort reflecting the clinical continuum of ADRD (see Table~\ref{tab:adrd_characteristics} for details). Following the established phenotype definitions~\cite{Zhang2025CrossDatasetDementiaProgression}, we stratify patients into three groups based on their diagnostic status: 
(1) \textbf{Cognitively Normal (CN):} Patients with no evidence of cognitive impairment. Following prior work~\cite{Li2023EarlyPredictionADRD_EHR}, we define the CN group as patients who were (1) aged above 40 years; (2) had at least one year of medical history in the database; (3) had no ADRD-related diagnoses, determined by excluding patients with any of the ICD codes defined below; and (4) had no exposure to dementia-related medications based on an established list~\cite{Albrecht2019AlgorithmDementiaSubtype}. These criteria yielded an initial CN pool of 91,264 patients. To mitigate class imbalance, we randomly downsampled the CN group at a $2\times$ ratio across three independent draws, resulting in a final CN sample of 11,408 discharge notes.
(2) \textbf{Mild Cognitive Impairment (MCI):} Patients in the symptomatic pre-dementia stage, characterized by subtle cognitive changes that do not significantly interfere with daily functioning. MCI status was defined using ICD-9 code \textit{331.83} and ICD-10 code \textit{G31.84}.
(3) \textbf{Alzheimer's Disease and Related Dementias (ADRD):} Patients with major neurocognitive disorder in which symptoms have progressed to impair independent functioning. ADRD status was defined using ICD codes for Alzheimer's disease (\textit{ICD-9: 331.0; ICD-10: G30.0, G30.1, G30.8, G30.9}), Lewy body dementia (\textit{ICD-9: 331.82; ICD-10: G31.83}), frontotemporal dementia (\textit{ICD-9: 331.1, 331.11, 331.19; ICD-10: G31.09}), vascular dementia (\textit{ICD-9: 290.40--290.43; ICD-10: F01.50, F01.51}), and non-specific dementias (\textit{ICD-9: 290.x, 294.x, 797; ICD-10: F02.80, F02.81, F03.90, F03.91}).

\header{Phenotype List.} We develop two phenotype lists and compare their results in experiments. Following prior work~\cite{Oh2023PhenotypesAD}, we first define \textit{Phenotype List 1}, which includes ten phenotypes organized into six broad categories: Memory Indicators, Comorbidities, Family History, Neurobehavioral Tests/Ratings, Neuroimaging Findings, and Biomarker Test Results. After consultation with ADRD experts, we then develop \textit{Phenotype List 2}, which contains 27 additional phenotypes spanning five categories: Memory, Executive Functions, Language, Visuospatial Skills, and Behavior. By comparing phenotype lists derived from different perspectives, we aim to provide a more comprehensive view of the domain. This strategy helps identify overlapping features while also highlighting differences between structured, data-driven phenotypes and those grounded in clinical expertise.

\begin{figure}[htbp!]
  \centering
    \includegraphics[width=0.8\textwidth]{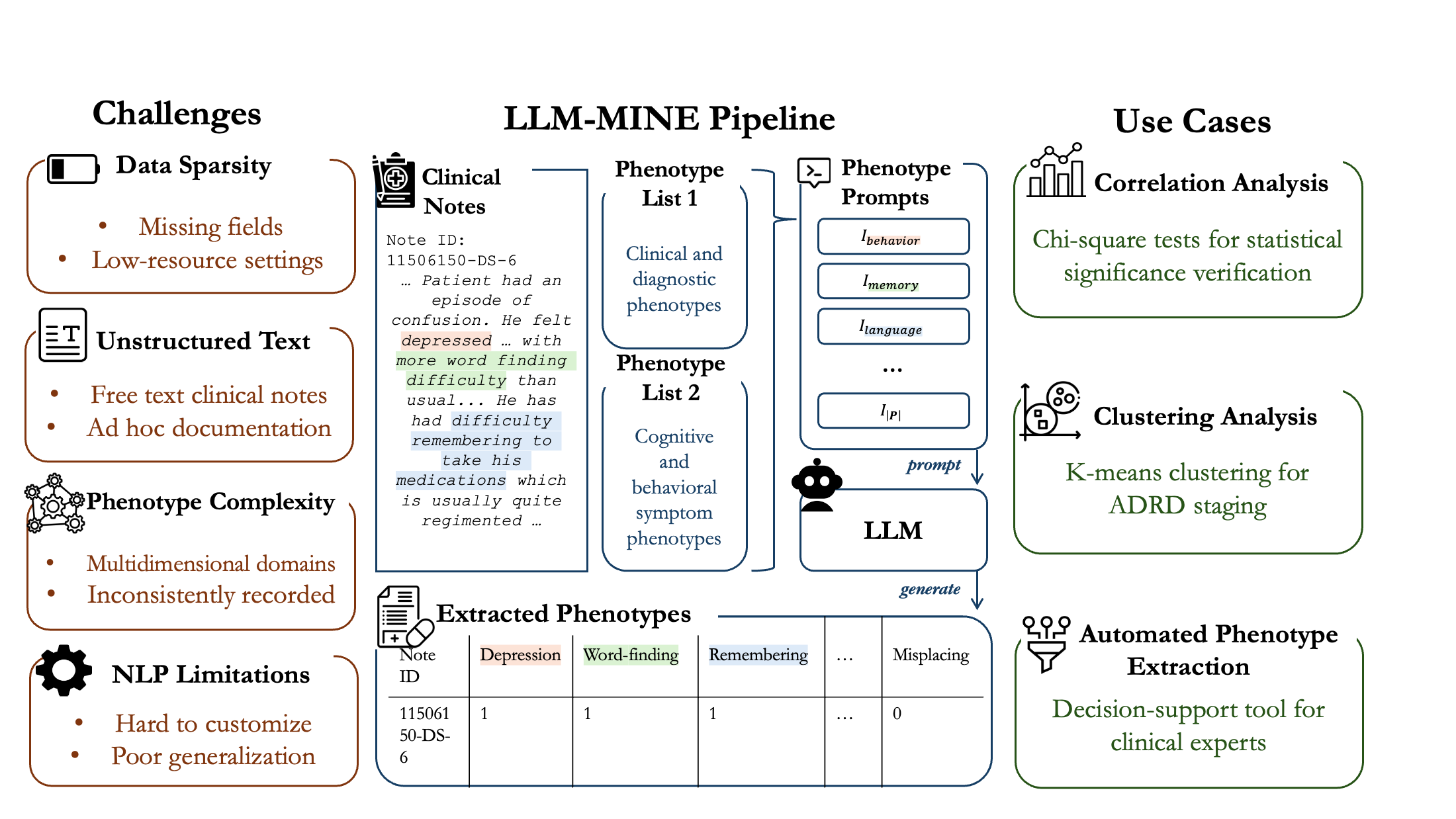}
 \caption{\textbf{Overview of the LLM-MINE framework.} \textit{(Left)} Key challenges motivating the use of LLMs for ADRD phenotype extraction from clinical notes. \textit{(Center)} The LLM-MINE pipeline: unstructured clinical notes are processed through expert-defined phenotype lists using prompting with a LLM to produce binary phenotype feature vectors. \textit{(Right)} Downstream use cases including statistical correlation analysis (chi-square tests), unsupervised disease staging (k-means clustering), and baseline comparison for automated phenotype extraction.}
 \label{fig:framework}
\end{figure}

\header{LLM-based Dementia Phenotype Mining Framework.}
\label{sec:methods sec1}
We propose \modelname \ (\textbf{LLM}-based de\textbf{M}ent\textbf{I}a phe\textbf{N}otyp\textbf{E} mining), a modular framework for extracting ADRD-related phenotypes from unstructured clinical notes. Figure~\ref{fig:framework} illustrates the overall pipeline. We use gemma-3-12b-it~\cite{gemmateam2025gemma3technicalreport}, an open-source instruction-tuned LLM whose parameters have been pre-trained on a massive amount of textual corpus including biomedical literature, public clinical databases, and general web data.
\modelname\ directly takes in raw clinical notes without any domain-specific preprocessing, tokenization, or concept normalization. Unlike traditional NLP pipelines that depend on handcrafted rules or tailored vocabulary mappings, \modelname\ leverages the broad medical knowledge encoded during pre-training to generalize across diverse note formats, abbreviations, and institution-specific clinical language. As the average length of the clinical notes in our dataset often exceeds the model's maximum context window, we partition each note into chunks at sentence boundaries, prompt the model to extract phenotypes from each chunk independently, and take the union of all extracted phenotypes across chunks to produce the final phenotype profile for that note. This chunking strategy ensures that no sentence is truncated mid-span, preserving local clinical context within each extraction call.

We frame phenotype extraction as a user-defined classification-style task rather than open-ended generation. For each phenotype category $\mathbf{P}_i$, the model is instructed to select only from a user-customizable candidate phenotypes defined in the phenotype lists, outputting a comma-separated list of present phenotypes or \texttt{`none'} if none are mentioned. This design choice reflects a deliberate trade-off between flexibility and controllability: while open generation would allow the model to capture phenotypic descriptions in its own terms, constraining the output space to a fixed vocabulary ensures consistent, structured results that can be directly mapped to binary feature vectors without post-processing. This approach also suppresses the model's tendency to produce explanatory reasoning that, while potentially useful for interpretability, introduces parsing complexity and increases the risk of hallucinations. While this comes at the cost of output interpretability, the constrained output form yields more accurate extractions, reduces output token cost, and facilitates downstream aggregation.

\modelname\ is a training-efficient pipeline and we use both zero-shot~\cite{kojima2023largelanguagemodelszeroshot} and few-shot prompting methods~\cite{brown2020languagemodelsfewshotlearners}. Zero-shot prompting instructs the model to perform extraction without providing task-specific examples, relying entirely on the model's pre-trained knowledge to interpret clinical language and map it to the target phenotype features. Few-shot prompting is used by appending three examples to each prompt. These demonstrations provide the model with concrete input-output pairs illustrating the expected extraction behavior, including cases where the correct output is \texttt{`none'}. This prompting method has been shown to increase task performance that is comparable to fine-tuning without the training costs~\cite{brown2020languagemodelsfewshotlearners}. Specifically, we use the following prompt templates:
\begin{tcolorbox}[promptbox, title={Zero-shot Prompt for Phenotype Extraction}]
You are analyzing a segment of a clinical nursing note. Extract the patient’s $[name(\mathbf{P}_i)]$ from the given discharge note. Please choose from $candidate([\mathbf{P}_i)]$. Return only the combination of the above outputs or \texttt{`none'} if none are mentioned in the note. \#\#Note\#\#: [$\mathbf{T}$].
\end{tcolorbox}

\begin{tcolorbox}[promptbox, title={Few-shot Prompt for Phenotype Extraction}]
You are analyzing a segment of a clinical nursing note. Extract the patient’s $[name(\mathbf{P}_i)]$ from the given discharge note. Please choose from $candidate([\mathbf{P}_i)]$. Return only the combination of the above outputs or \texttt{`none'} if none are mentioned in the note. 
Examples: $[(\mathbf{E}_i)]$
\#\#Note\#\#: [$\mathbf{T}$].
\end{tcolorbox}

\noindent In the prompt, text in ``[]'' are placeholders to be filled with specific values. For instance, the filled prompt for comorbidities is ``\textit{You are analyzing a segment of a clinical nursing note. Extract the patient’s comorbidities of ADRD from the given discharge note. Please choose from `hypertension' and `depression'. Return only the combination of the above outputs or \texttt{`none'} if none are mentioned in the note.}''

\section*{Experimental Results}
In this section, we present a series of experiments designed to evaluate the effectiveness and clinical utility of our proposed \modelname. We organize our evaluation around three research questions addressing phenotype extraction quality, downstream discriminative power, and comparison against established baselines.


\header{RQ1: What is the clinical utility for LLM-based automated phenotyping of patients on clinical notes?}
We begin by examining the accuracy of our LLM-based phenotype extraction approach. We extract phenotypes following the definitions in both phenotype lists from clinical notes. Because the MCI cohort contains only 992 notes, we randomly sample 1{,}000 notes from each of the CN and ADRD groups to maintain cohort balance. The extraction results, indicated by phenotype presence per cohort, are presented in Table~\ref{tab:phenotype1} (Phenotype List~1) and Table~\ref{tab:phenotype2} (Phenotype List~2). For each phenotype, we calculate the number of occurrences; instances containing none of the phenotypes in a given category are marked as ``None''.
From the extraction result, we observe several noteworthy findings. 
\begin{wraptable}{r}{0.50\linewidth}
\centering
\vspace{0.2cm}
\caption{\textbf{LLM phenotype mining result for Phenotype List 1.}}
\vspace{-5pt}
\label{tab:phenotype1}
\resizebox{0.95\linewidth}{!}{%
\begin{tabular}{ccccc}
        \toprule
        Category & Phenotype & CN & MCI & ADRD \\
        \midrule
        \multirow{3}{*}{Memory Indicators}
        & None & 514 & 275 & 336 \\
        & Repeating & 476 & 698 & 652 \\
        & Misplacing & 311 & 582 & 548 \\
        \midrule
        \multirow{3}{*}{Comorbidities}
        & None & 39 & 27 & 16 \\
        & Hypertension & 912 & 896 & 955 \\
        & Depression & 665 & 785 & 860 \\
        \midrule
        \multirow{2}{*}{Family history}
        & None & 490 & 494 & 589 \\
        & Family history & 510 & 498 & 411 \\
        \midrule
        \multirow{3}{*}{Neurobehavioral tests}
        & None & 986 & 947 & 952 \\
        & MMSE & 13 & 42 & 34 \\
        & CDR & 1 & 10 & 7 \\
        \midrule
        \multirow{3}{*}{Neuroimaging findings}
        & None & 677 & 498 & 399 \\
        & Atrophy & 238 & 413 & 521 \\
        & Infarct & 303 & 457 & 564 \\
        \midrule
        \multirow{2}{*}{Biomarker test results}
        & None & 997 & 980 & 978 \\
        & tau & 3 & 12 & 22 \\
        \bottomrule
      \end{tabular}
}
\end{wraptable} 
First, most phenotypes \textbf{show a monotonic increase in prevalence from the CN group, through MCI, to the ADRD group}, particularly in Phenotype List~2. 
A small number of categories deviate from this pattern, including memory indicators and family history in Phenotype List~1, and behavior in Phenotype List~2.
Phenotype List~2 yields a \textbf{higher overall phenotype prevalence than Phenotype List~1}, largely because List~2 targets cognitive-domain symptoms that are near-universal among impaired patients. For instance, in the Memory category, fewer than 4\% of MCI and ADRD patients have ``None'' recorded (MCI: 32/992, ADRD: 17/1{,}000), whereas in List~1, 27--34\% of impaired patients show no memory indicators. This suggests that List~2's cognitive-domain phenotypes are more consistently documented in clinical notes and offer broader coverage of disease-relevant features. Memory symptoms are near-universal among MCI and ADRD patients but substantially less common in CNs. For instance, ``Recent Events'' memory problems appear in 926 of 992 MCI and 973 of 1{,}000 ADRD patients, compared with only 648 of 1{,}000 CNs. 
Similarly, ``Missing Appointments'' is documented in 941 MCI and 976 ADRD patients versus 799 CNs. These findings confirm that \textbf{memory impairment is the strongest discriminator between cohorts}.
Some phenotypes display \textbf{clinically interpretable non-monotonic patterns}. ``Driving problem'' decreases across CN (222), MCI (193), and ADRD (113). Although counterintuitive at first glance, this likely reflects the fact that patients with severe dementia are no longer driving, causing driving-related concerns to disappear from their clinical notes entirely. Additionally, executive function and language problems sharply separate MCI and ADRD from CNs but \textbf{offer limited discrimination} between MCI and ADRD. For example, ``Judgement'' is documented in 609 CNs but jumps to 943 in MCI and 967 in ADRD---a large increase from CN to MCI with negligible further change. This suggests that executive and language features are effective for detecting cognitive impairment but less useful for staging severity.

We further perform chi-square tests of independence~\cite{McHugh2013ChiSquareIndependence} to measure the statistical significance of differences between the cohorts, which is presented in Table~\ref{tab:chi_square_dataset_a} and Table~\ref{tab:chi_square_dataset_b}. For each phenotype, we construct a $2 \times 3$ contingency table comparing the presence/absence of the phenotype across the three cohorts. Pairwise $2\times 2$ tests are subsequently conducted between each cohort pair (CN vs. MCI, CN vs. ADRD, \textit{etc}). Statistical significance was defined as $p < 0.05$.
\begin{wraptable}{r}{0.50\linewidth}
\centering
\caption{\textbf{LLM phenotype mining result for Phenotype List 2.}}
\vspace{-5pt}
\label{tab:phenotype2}
\resizebox{0.95\linewidth}{!}{%
      \begin{tabular}{ccccc}
        \toprule
        Category & Phenotype & CN & MCI & ADRD \\
        \midrule
        \multirow{5}{*}{Memory}
        & None & 189 & 32 & 17 \\
        & Recent Events & 648 & 926 & 973 \\
        & Remote Events & 116 & 254 & 460 \\
        & Misplacing & 654 & 918 & 959 \\
        & Missing Appointments & 799 & 941 & 976 \\
        \midrule
        \multirow{6}{*}{Executive Functions}
        & None & 233 & 43 & 31 \\
        & Planning and organization & 206 & 618 & 611 \\
        & Multi-tasking & 17 & 41 & 22 \\
        & Concentration & 163 & 529 & 388 \\
        & Judgement & 609 & 943 & 967 \\
        & Problem-solving & 519 & 908 & 933 \\
        \midrule
        \multirow{6}{*}{Language}
        & None & 679 & 318 & 284 \\
        & Word-finding & 55 & 330 & 230 \\
        & Slurred speech & 77 & 150 & 161 \\
        & Halting or stuttering & 29 & 87 & 103 \\
        & Impairment & 60 & 259 & 231 \\
        & Abnormal speech & 313 & 664 & 706 \\
        \midrule
        \multirow{6}{*}{Visuospatial Skills}
        & None & 216 & 41 & 25 \\
        & Finding way around & 154 & 287 & 280 \\
        & Lost in familiar places & 82 & 255 & 320 \\
        & Driving problem & 222 & 193 & 113 \\
        & Seeing things & 713 & 894 & 912 \\
        & Recognizing objects or faces & 536 & 894 & 957 \\
        \midrule
        \multirow{9}{*}{Behavior}
        & None & 134 & 45 & 53 \\
        & Emotional expression & 254 & 566 & 607 \\
        & Personality or behavior & 277 & 693 & 742 \\
        & Agitation and aggression & 107 & 242 & 290 \\
        & Apathy or decreased motivation & 265 & 636 & 762 \\
        & Hygiene and eating & 481 & 539 & 585 \\
        & Depression or anxiety & 606 & 771 & 791 \\
        & Hallucination & 224 & 582 & 707 \\
        & Weight change & 846 & 920 & 915 \\
        \bottomrule
      \end{tabular}
}
\end{wraptable}
Several key findings emerge from the chi-square analysis. \textbf{First, the MCI vs.\ ADRD comparison is consistently the weakest across both phenotype lists}. In List~2, only Memory reaches statistical significance for this pairwise comparison ($p < 0.05$), while Behavior is essentially non-significant ($p = 0.494$) and Visuospatial Skills ($p = 0.056$) and Language ($p = 0.084$) remain at borderline significance. This confirms that while LLM-extracted phenotypes are effective at detecting cognitive impairment, they have limited power for distinguishing disease stage.
\textbf{Second, several individual patterns merit attention}. Family History in List~1 exhibits an unusual profile: it is highly significant overall and for CN vs.\ ADRD and MCI vs.\ ADRD ($p < 0.001$), yet non-significant for CN vs.\ MCI ($p = 0.755$). This suggests that family history documentation does not differ between CNs and MCI patients but becomes more thoroughly recorded as disease progresses and patients undergo more comprehensive clinical workups. Comorbidities is the weakest discriminator in List~1, achieving only $p = 0.007$ overall, with a non-significant CN vs.\ MCI comparison ($p = 0.179$). This is consistent with the high baseline prevalence of conditions such as hypertension and depression across all groups, which diminishes their discriminative utility. By contrast, neuroimaging findings is the only List~1 category that achieves $p < 0.001$ across all comparisons, including MCI vs.\ ADRD, making it the single strongest discriminator for three-way group separation.
\textbf{Finally, List~2 phenotypes are more uniformly significant than those in List~1}. All five List~2 categories achieve $p < 0.001$ for the overall, CN vs.\ MCI, and CN vs.\ ADRD comparisons, whereas List~1 contains categories with weaker or non-significant pairwise results. This suggests that cognitive-domain phenotypes (memory, executive function, language) are more consistently informative than the clinical and demographic features captured in List~1.


\begin{table}[htbp!]
  \centering
  \caption{\textbf{Chi-square Test P-values Results.} Significance levels: *** $p<0.001$, ** $p<0.01$, * $p<0.05$.}
  \begin{subtable}[t]{0.48\linewidth}
  \caption{ Phenotype List 1.}\label{tab:chi_square_dataset_a}
  \resizebox{0.95\linewidth}{!}{%
  \begin{tabular}{lcccc}
        \toprule
        Phenotypes & Overall & CN vs. MCI & CN vs. ADRD & MCI vs. ADRD \\
        \midrule
        Memory Indicators          & *** & *** & *** & ** \\
        Comorbidities              & **              & $0.179$              & **              & $0.117$ \\
        Family History             & *** & $0.755$              & *** & *** \\
        Neurobehavioral tests & *** & *** & *** & $0.863$ \\
        Neuroimaging findings      & *** & *** & *** & *** \\
        Biomarker test results     & *** & *              & *** & $0.125$ \\
        \bottomrule
      \end{tabular}
      }
      \end{subtable}
      \hfill
      \begin{subtable}[t]{0.48\linewidth}
      \caption{Phenotype List 2.}\label{tab:chi_square_dataset_b}
      \resizebox{0.98\linewidth}{!}{%
      \begin{tabular}{lcccc}
        \toprule
        Phenotypes & Overall & CN vs. MCI & CN vs. ADRD & MCI vs. ADRD \\
        \midrule
        Memory              & *** & *** & *** & * \\
        Executive Functions & *** & *** & *** & $0.181$ \\
        Language            & *** & *** & *** & $0.084$ \\
        Visuospatial skills & *** & *** & *** & $0.056$ \\
        Behavior            & *** & *** & *** & $0.494$ \\
        \bottomrule
      \end{tabular}
      }
      \end{subtable}
\end{table}

\header{RQ2: What diagnostic discriminative power do the resulting phenotypic profiles hold when used to stratify patients?}
With the ADRD-centric phenotypes extracted from clinical notes, we perform K-means clustering with $K=2$ and $K=3$. Each patient is represented as a feature vector consisting of binary phenotype indicators (1 = present, 0 = not present) across the extracted phenotype categories. We use Euclidean distance as the distance metric. 
For $K=2$, we organize the cohort into CN and patient, with the patient group consisting of both MCI and ADRD patients from the original cohorts. We do this because of the earlier observation from the chi-square tests that MCI against ADRD patients in general has higher p-values across phenotypes, suggesting their distributions are not statistically significantly different. Since treating two highly similar groups as distinct clusters may hinder clustering performance, we collapse them into a single patient group and evaluate whether reducing K from 3 to 2 yields better clustering results. For $K=3$, the group labels remain the same as the original cohort. 
\begin{wraptable}{r}{0.50\linewidth}
\centering
\vspace{0.3cm}
\caption{\textbf{Clustering results for compared methods}.}
\vspace{-5pt}
\label{tab:clustering}
\resizebox{0.95\linewidth}{!}{%
\begin{tabular}{ccccc}
\toprule
Phenotype & Setting & ARI & NMI & FMI \\
\midrule
\multirow{2}{*}{QuickUMLS} 
& n = 2 &  0.004 & 0.000 & 0.657\\
& n = 3 & 0.003 & 0.003 & 0.378  \\
\hline
\multirow{2}{*}{BioNER} 
& n = 2 & 0.003 & 0.006 & 0.530 \\
& n = 3 & 0.011 & 0.011 & 0.341 \\
\hline
\multirow{4}{*}{Phenotype List 1} 
& n = 2 w/ zero-shot & -0.001 & 0.000 & 0.546 \\
& n = 3 w/ zero-shot & 0.034 & 0.032
& 0.358 \\
& n = 2 w/ few-shot & 0.052 & 0.023 & 0.568  \\
& n = 3 w/ few-shot &  0.030 & 0.029 & 0.363 \\
\hline
\multirow{4}{*}{Phenotype List 2} 
& n = 2 w/ zero-shot & 0.190 & 0.120 & 0.628 \\
& n = 3 w/ zero-shot & 0.119 &  0.105 & 0.483 \\
& n = 2 w/ few-shot & 0.277 & 0.226 & 0.660   \\
& n = 3 w/ few-shot & 0.172 & 0.166 & 0.448 \\
\hline
\multirow{2}{*}{Phenotype List 1 \& 2} 
& n = 2 w/ few-shot  & \textbf{0.290} & \textbf{0.232} & \textbf{0.666} \\
& n = 3 w/ few-shot  & 0.168 & 0.165 & 0.446 \\ 
\bottomrule
\end{tabular}
}
\end{wraptable} 
By this we explore more fine-grained clustering and see if our extracted phenotype features are distinct enough to correctly cluster the patients. Clustering performance was evaluated by comparing the resulting cluster assignments with cohort labels using external validation metrics, including Adjusted Rand Index (ARI), Normalized Mutual Information (NMI), and Fowlkes–Mallows Index (FMI). The clustering results for both phenotype lists, using both zero-shot and few-shot prompting methods, are presented in Table~\ref{tab:clustering}. 
From the result, we observe a \textbf{lower performance across Phenotype List~1 compared to Phenotype List~2}, with the highest performing set being $K=2$ clustering using features extracted with few-shot prompting. \textbf{Few-shot prompting consistently improves performance over zero-shot across both phenotype lists}. For Phenotype List~2, few-shot prompting improves ARI by 46\% ($0.190 \rightarrow 0.277$) and NMI by 88\% ($0.120 \rightarrow 0.226$) under $K=2$, indicating that more accurate phenotype extraction directly translates to stronger clustering signal. We also observe \textbf{consistently higher performance for $\bm{K=2}$ over $\bm{K=3}$} for Phenotype List~2, consistent with the chi-square finding that MCI and ADRD phenotype distributions are not statistically significantly different, making the two-group partition easier for the clustering algorithm to recover.

\begin{figure}[H]
  \centering
  \begin{subfigure}[t]{0.40\textwidth}
    \centering
    \includegraphics[width=\textwidth]{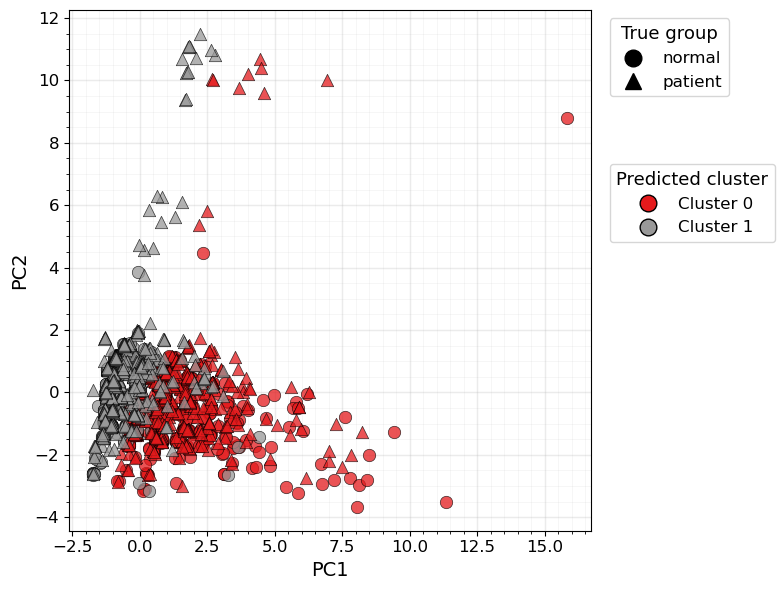}
    \caption{PCA for n=2 clustering.}
    \label{fig:phe1n2}
  \end{subfigure}
  \hfill
  \begin{subfigure}[t]{0.40\textwidth}
    \centering
    \includegraphics[width=\textwidth]{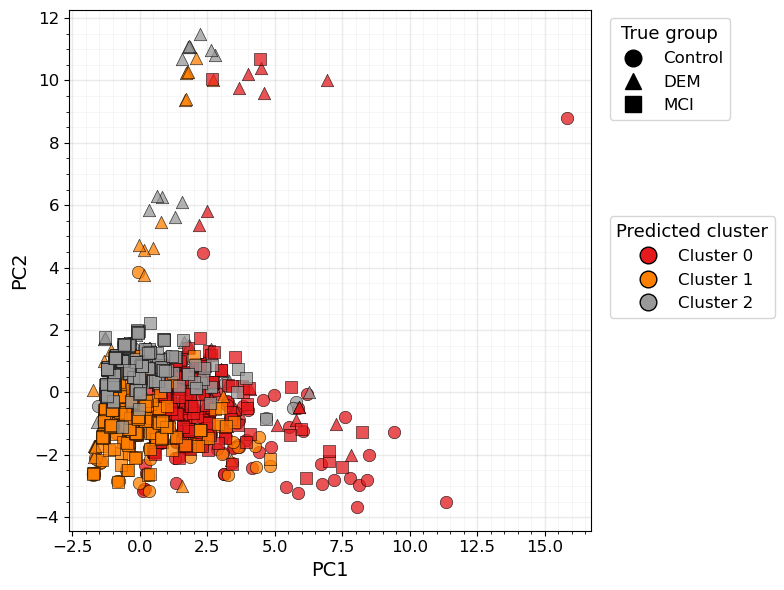}
    \caption{PCA for n=3 clustering.}
    \label{fig:phe1n3}
  \end{subfigure}

  \caption{PCA for Phenotype List~1. Phenotypes are extracted using few-shot prompting method.}
  \label{fig:combined1}
\end{figure}

\begin{figure}[H]
  \centering
  \begin{subfigure}[t]{0.40\textwidth}
    \centering
    \includegraphics[width=\textwidth]{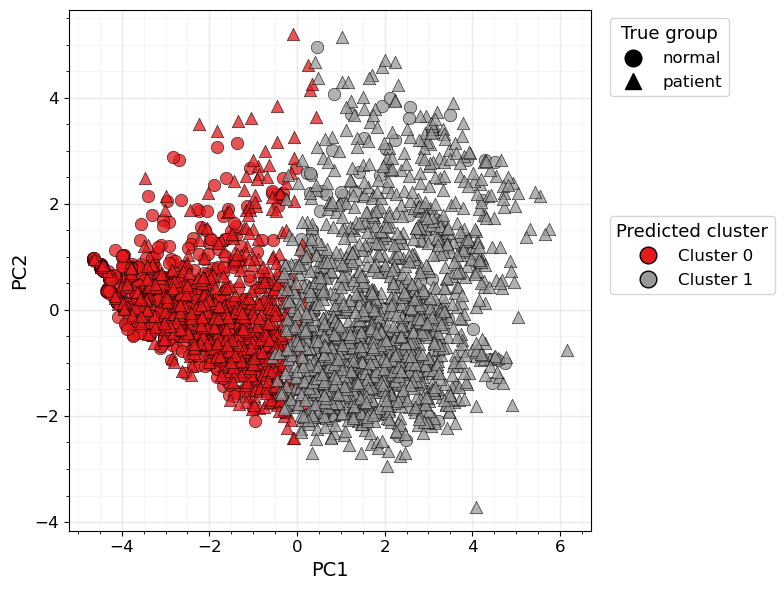}
    \caption{PCA for n=2 clustering.}
    \label{fig:phe2n2}
  \end{subfigure}
  \hfill
  \begin{subfigure}[t]{0.40\textwidth}
    \centering
    \includegraphics[width=\textwidth]{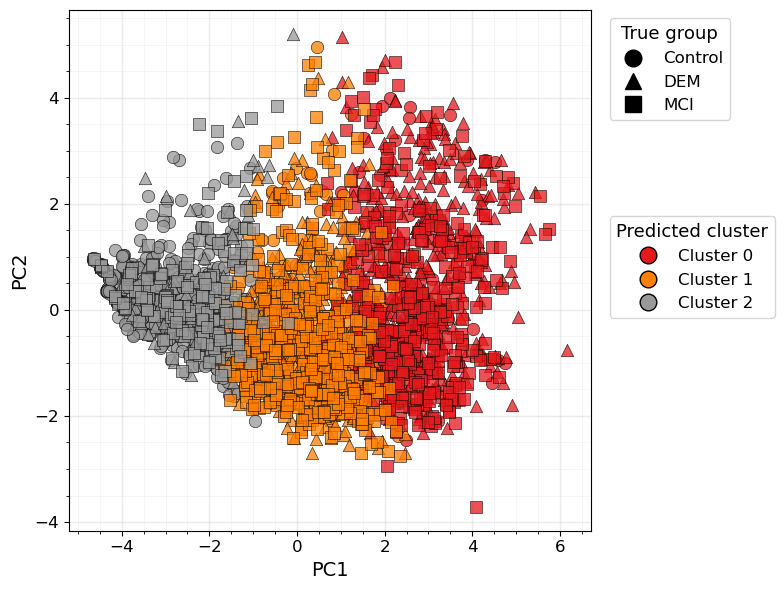}
    \caption{PCA for n=3 clustering.}
    \label{fig:phe2n3}
  \end{subfigure}

  \caption{PCA for Phenotype List~2. Phenotypes are extracted using few-shot prompting method.}
  \label{fig:combined2}
\end{figure}

To further understand the structure of the extracted features, we perform principal component analysis (PCA) and visualize the patient representations in two-dimensional space. The plots are presented in Figure~\ref{fig:combined1} and Figure~\ref{fig:combined2}. The PCA plots further verify the clustering performance. We observe that the true groups overlap substantially in 2-dimensional PCA space, suggesting the clusters are not well-separated with compact, spherical geometry in the feature space, which is consistent with the low performance of k-means clustering. Overall, the PCA results confirm that the patient stratification task is inherently difficult: the phenotype distributions do not form compact, well-separated clusters, which limits the performance ceiling of distance-based methods such as K-means. 

\header{RQ3: Is our LLM-based phenotype extraction approach better than established approaches?}
We compare the performance of our proposed method with established methods. One baseline method is to utilize biomedical-ner-all~\cite{raza2022large}, a transformer model trained on annotated biomedical data, to extract all comorbidities and symptoms as phenotypes. We also use a dictionary based method quickUMLS~\cite{Soldaini2016QuickUMLSAF} that matches similar biomedical patterns in the Unified Medical Language System (UMLS) metathesaurus. For biomedical-ner-all, we keep only the concepts with a confidence score greater than or equal to 0.8 to ensure extraction accuracy. For quickUMLS, we restrict the extraction to words longer than 4 characters, and ones that appear in at least 50 notes to filter out rare and noisy concepts. We then perform $K=2$ and $K=3$ clustering based on the one hot encoded concepts. The baseline clustering results, alongside our LLM-based extraction clustering results, are presented in Table~\ref{tab:clustering}. 

The comparison result reflects the advantageous performance of our framework compared to the two baseline methods. Overall performance remains low in absolute terms across all methods, reflecting the difficulty of the stratification task. Among evaluation metrics, ARI and NMI are the most informative, as FMI is heavily influenced by cluster size imbalance and therefore less sensitive to meaningful structural differences between cluster assignments; we report it for completeness but focus our analysis on ARI and NMI.
Under the same $K=2$ setting, our best method, combining Phenotype List~1 and Phenotype List~2 with few-shot prompting, improves ARI from 0.011 to 0.290 ($\Delta= +0.279$), NMI from 0.011 to 0.232 ($\Delta= +0.221$), and FMI from 0.657 to 0.666 ($\Delta= +0.009$). Our results suggest that the current baselines are not competitive in this clinical stratification setting, while our approach captures substantially more structured and disease-relevant information.

\begin{wraptable}{r}{0.55\linewidth}
\centering
\caption{\textbf{Human evaluation of LLM-extracted phenotypes for Phenotype List 1.} 
Each annotator column reports the consistency between that annotator's labels and the LLM output.}
\vspace{-5pt}
\label{tab:humaneval}
\resizebox{0.95\linewidth}{!}{%
      \begin{tabular}{ccccc}
        \toprule
        Phenotype &
        Ann. 1 & Ann. 2 & Ann. 3 & Fleiss' kappa \\
        \midrule
        Memory Indicators
        & 1.00 & 1.00 & 1.00 & 1.00 \\
        Comorbidities 
        & 1.00 & 1.00 & 1.00 & 1.00$^{\dagger}$  \\
        Family History 
        & 0.70 & 0.60 & 0.60 & 0.64 \\
        Neurobehavioral Tests 
        & 0.90 & 1.00 & 1.00 & 0.93$^{\dagger}$  \\
        Neuroimaging Findings 
        & 0.75 & 0.60 & 0.70 & 0.66 \\
        Biomarker Test Results 
        & 1.00 & 1.00 & 1.00 & 1.00$^{\dagger}$  \\
        \bottomrule
        \multicolumn{5}{l}{\footnotesize $^{\dagger}$Percent agreement is reported because Fleiss' kappa was undefined due to no label variation.}
      \end{tabular}
}
\end{wraptable}
We also conduct a \textbf{\textit{human evaluation}} to verify the accuracy of the LLM-extracted phenotypes. Three human annotators independently review 20 sampled notes, and the results are shown in Table~\ref{tab:humaneval}. We measure inter-annotator agreement using Fleiss’ kappa~\cite{fleiss1971measuring} for each binary phenotype category across the three annotators. For categories with no label variation, including comorbidities, neurobehavioral tests/ratings, and biomarker test results, Fleiss’ kappa is undefined; therefore, we report percent agreement instead.
Overall, the LLM-extracted results show high consistency with human annotations for memory indicators, comorbidities, neurobehavioral tests/ratings, and biomarker test results. However, agreement is lower for family history and neuroimaging findings. Error analysis shows that these lower scores are mainly caused by overly broad interpretations by the LLM. For family history, the LLM sometimes extracts family history of any disease, even though the prompt defines this category as ADRD-related family history only. This leads to several false positive cases. Similarly, for neuroimaging findings, the LLM sometimes identifies atrophy or infarct in any organ system, rather than restricting extraction to brain-related atrophy or cerebral infarct.

\section*{Conclusion}
We present \modelname, a modular, user-customizable, and training-free framework that leverages an LLM to extract ADRD-related phenotypes directly from clinical notes without domain-specific preprocessing or model fine-tuning. Through systematic evaluation on a real-world cohort, we verify that the extracted phenotypes exhibit statistically significant differences across the CN, MCI, and ADRD groups, and demonstrate substantial improvements over dictionary-based and biomedical NER baselines in unsupervised disease staging. Our analyses further reveal that cognitive and behavioral symptom phenotypes provide more consistently informative signals than clinical and diagnostic markers, and that distinguishing MCI from ADRD remains a fundamental challenge. 
These findings highlight the potential practical value of our LLM-based framework for scalable phenotype extraction, early risk stratification, and data-driven dementia research using unstructured clinical documentation. At the same time, the performance of our framework is inherently shaped by the general-purpose nature of the underlying language model. With the recent emergence of domain-adapted medical LLMs, such as Google's MedGemma~\cite{sellergren2025medgemma}, our future plan is to evaluate whether models with clinical pretraining yield more precise phenotype extraction.

\header{Limitations.}
There are several limitations of our work that warrant discussion. First, we used ICD diagnostic codes as a proxy for clinical cognitive status. Because coding practices vary across providers and institutions, and because codes assigned during acute encounters may not reflect a patient's underlying cognitive trajectory, this proxy introduces potential misclassification that could attenuate or distort group-level differences. Second, although LLMs provide a scalable approach for extracting phenotypes from unstructured clinical notes, their performance may be affected by limited interpretability, potential bias, and sensitivity to prompt design as reflected in the human evaluation. 

\header{Ethical Considerations.} This research involved secondary analysis of a publicly available, de-identified dataset. As the study did not involve interaction with human subjects or access to identifiable private information, it was deemed Not Human Subject Research by the Institutional Review Board (IRB).

\header{Acknowledgments.}
We thank Dr. Andrew Breithaupt for providing clinical guidance on ADRD phenotype definitions and validation.
This research was partially supported by the internal funds and GPU servers provided by  School of Nursing’s Center
for Data Science of Emory University.

\makeatletter
\renewcommand{\@biblabel}[1]{\hfill #1.}
\makeatother

\bibliographystyle{vancouver}
\bibliography{amia}  

@article{Oh2023PhenotypesAD,
  author  = {Oh, I. Y. and Schindler, S. E. and Ghoshal, N. and Lai, A. M. and Payne, P. R. O. and Gupta, A.},
  title   = {Extraction of clinical phenotypes for Alzheimer's disease dementia from clinical notes using natural language processing},
  journal = {JAMIA Open},
  year    = {2023},
  volume  = {6},
  number  = {1},
  pages   = {ooad014},
  date    = {2023-02-24},
  doi     = {10.1093/jamiaopen/ooad014},
  pmid    = {36844369},
  pmcid   = {PMC9952043}
}

@article{2023MIMIC,
  title={MIMIC-IV, a freely accessible electronic health record dataset},
  author={ Johnson, Alistair E. W.  and  Bulgarelli, Lucas  and  Shen, Lu  and  Gayles, Alvin  and  Shammout, Ayad  and  Horng, Steven  and  Pollard, Tom J.  and  Moody, Benjamin  and  Gow, Brian  and  Lehman, Li Wei H. },
  journal={Scientific Data},
  volume={10},
  number={1},
  year={2023},
}

@misc{WHO_Dementia,
    title        = {Dementia},
    author       = {{World Health Organization}},
    year         = 2025,
    month        = mar,
    howpublished = {\url{https://www.who.int/en/news-room/fact-sheets/detail/dementia}},
    note         = {Accessed: 2026-01-25},
}

@article{alzheimer2025,
  title={2025 Alzheimer's disease facts and figures},
  author={Alzheimer's Association},
  journal={Alzheimer's \& Dementia},
  volume={21},
  number={5},
  year={2025},
  publisher={Wiley Online Library}
}

@article{lyketsos2022standardizing,
  title={Standardizing electronic health record data on AD/ADRD to accelerate health equity in prevention, detection, and treatment},
  author={Lyketsos, CG and Roberts, SB and Swift, Elaine K and Quina, A and Moon, G and Kremer, I and Tariot, P and Fillit, H and Bovenkamp, DE and Zandi, PP and others},
  journal={The journal of prevention of Alzheimer's disease},
  volume={9},
  number={3},
  pages={556--560},
  year={2022},
  publisher={Springer}
}

@inproceedings{xie2025hypkg,
  title={HypKG: Hypergraph-Based Knowledge Graph Contextualization for Precision Healthcare},
  author={Xie, Yuzhang and Han, Xu and Xu, Ran and Hu, Xiao and Lu, Jiaying and Yang, Carl},
  booktitle={International Semantic Web Conference},
  pages={328--348},
  year={2025},
  organization={Springer}
}

@article{chen2023assess,
  title={Assess the documentation of cognitive tests and biomarkers in electronic health records via natural language processing for Alzheimer’s disease and related dementias},
  author={Chen, Zhaoyi and Zhang, Hansi and Yang, Xi and Wu, Songzi and He, Xing and Xu, Jie and Guo, Jingchuan and Prosperi, Mattia and Wang, Fei and Xu, Hua and others},
  journal={International journal of medical informatics},
  volume={170},
  pages={104973},
  year={2023},
  publisher={Elsevier}
}

@article{cheng2025high,
  title={High-Throughput Phenotyping of the Symptoms of Alzheimer Disease and Related Dementias Using Large Language Models: Cross-Sectional Study},
  author={Cheng, You and Malekar, Mrunal and He, Yingnan and Bommareddy, Apoorva and Magdamo, Colin and Singh, Arjun and Westover, Brandon and Mukerji, Shibani S and Dickson, John and Das, Sudeshna},
  journal={JMIR AI},
  volume={4},
  pages={e66926},
  year={2025},
  publisher={JMIR Publications Toronto, Canada}
}

@misc{kojima2023largelanguagemodelszeroshot,
      title={Large Language Models are Zero-Shot Reasoners}, 
      author={Takeshi Kojima and Shixiang Shane Gu and Machel Reid and Yutaka Matsuo and Yusuke Iwasawa},
      year={2023},
      eprint={2205.11916},
      archivePrefix={arXiv},
      primaryClass={cs.CL},
      url={https://arxiv.org/abs/2205.11916}, 
}

@misc{brown2020languagemodelsfewshotlearners,
      title={Language Models are Few-Shot Learners}, 
      author={Tom B. Brown and Benjamin Mann and Nick Ryder and Melanie Subbiah and Jared Kaplan and Prafulla Dhariwal and Arvind Neelakantan and Pranav Shyam and Girish Sastry and Amanda Askell and Sandhini Agarwal and Ariel Herbert-Voss and Gretchen Krueger and Tom Henighan and Rewon Child and Aditya Ramesh and Daniel M. Ziegler and Jeffrey Wu and Clemens Winter and Christopher Hesse and Mark Chen and Eric Sigler and Mateusz Litwin and Scott Gray and Benjamin Chess and Jack Clark and Christopher Berner and Sam McCandlish and Alec Radford and Ilya Sutskever and Dario Amodei},
      year={2020},
      eprint={2005.14165},
      archivePrefix={arXiv},
      primaryClass={cs.CL},
      url={https://arxiv.org/abs/2005.14165}, 
}

@article{raza2022large,
  title={Large-scale application of named entity recognition to biomedicine and epidemiology},
  author={Raza, Shaina and Reji, Deepak John and Shajan, Femi and Bashir, Syed Raza},
  journal={PLOS Digital Health},
  volume={1},
  number={12},
  pages={e0000152},
  year={2022},
  publisher={Public Library of Science San Francisco, CA USA}
}

@inproceedings{Soldaini2016QuickUMLSAF,
  title={QuickUMLS: a fast, unsupervised approach for medical concept extraction},
  author={Luca Soldaini},
  year={2016},
  url={https://api.semanticscholar.org/CorpusID:2990304}
}

@article{Huckvale2019DigitalPhenotyping,
  author  = {Huckvale, Kit and Venkatesh, Svetha and Christensen, Helen},
  title   = {Toward clinical digital phenotyping: a timely opportunity to consider purpose, quality, and safety},
  journal = {npj Digital Medicine},
  year    = {2019},
  volume  = {2},
  pages   = {88},
  doi     = {10.1038/s41746-019-0166-1}
}

@article{Bilal2025NLP_CancerEHR_Review,
  author  = {Bilal, Muhammad and Hamza, Ameer and Malik, Nadia},
  title   = {NLP for Analyzing Electronic Health Records and Clinical Notes in Cancer Research: A Review},
  journal = {Journal of Pain and Symptom Management},
  year    = {2025},
  volume  = {69},
  number  = {5},
  pages   = {e374--e394},
  issn    = {0885-3924},
  doi     = {10.1016/j.jpainsymman.2025.01.019}
}

@article{Ruckdeschel2023SmokingHistoryEHR,
  author  = {Ruckdeschel, J. and Riley, M. and Parsatharathy, S. and Chamarthi, R. and Rajagopal, C. and Hsu, H. and Mangold, D. and Driscoll, C.},
  title   = {Unstructured Data Are Superior to Structured Data for Eliciting Quantitative Smoking History From the Electronic Health Record},
  journal = {JCO Clinical Cancer Informatics},
  year    = {2023},
  volume  = {7},
  doi     = {10.1200/CCI.22.00155}
}

@misc{Shao_2025,
      title={LLM-MINE: Large Language Model based Alzheimer's Disease and Related Dementias Phenotypes Mining from Clinical Notes}, 
      author={Mingchen Shao and Yuzhang Xie and Carl Yang and Jiaying Lu},
      year={2026},
      eprint={2603.13673},
      archivePrefix={arXiv},
      primaryClass={cs.AI},
      url={https://arxiv.org/abs/2603.13673}, 
}

@inproceedings{xie2024promptlink,
  title={Promptlink: leveraging large language models for cross-source biomedical concept linking},
  author={Xie, Yuzhang and Lu, Jiaying and Ho, Joyce and Nahab, Fadi and Hu, Xiao and Yang, Carl},
  booktitle={Proceedings of the 47th International ACM SIGIR Conference on Research and Development in Information Retrieval},
  pages={2589--2593},
  year={2024}
}

@article{Walling2023DementiaEHRPhenotypes,
  author  = {Walling, A. M. and Pevnick, J. and Bennett, A. V. and Vydiswaran, V. G. V. and Ritchie, C. S.},
  title   = {Dementia and electronic health record phenotypes: a scoping review of available phenotypes and opportunities for future research},
  journal = {Journal of the American Medical Informatics Association},
  year    = {2023},
  volume  = {30},
  number  = {7},
  pages   = {1333--1348},
  date    = {2023-06-20},
  doi     = {10.1093/jamia/ocad086},
  pmid    = {37252836},
  pmcid   = {PMC10280354}
}

@article{Zhang2025CrossDatasetDementiaProgression,
  author  = {Zhang, C. and An, L. and Wulan, N. and Nguyen, K. N. and Orban, C. and Chen, P. and Chen, C. and Zhou, J. H. and Liu, K. and Yeo, B. T. T. and {Alzheimer's Disease Neuroimaging Initiative} and {Australian Imaging Biomarkers and Lifestyle Study of Aging}},
  title   = {Cross-Dataset Evaluation of Dementia Longitudinal Progression Prediction Models},
  journal = {Human Brain Mapping},
  year    = {2025},
  volume  = {46},
  number  = {11},
  pages   = {e70280},
  date    = {2025-08-01},
  doi     = {10.1002/hbm.70280},
  pmid    = {40748187},
  pmcid   = {PMC12315237}
}

@article{Sperling2011PreclinicalAD,
  author  = {Sperling, Reisa A. and Aisen, Paul S. and Beckett, Laurel A. and Bennett, David A. and Craft, Suzanne and Fagan, Anne M. and Iwatsubo, Takeshi and Jack, Clifford R., Jr. and Kaye, Jeffrey and Montine, Thomas J. and Park, Denise C. and Reiman, Eric M. and Rowe, Christopher C. and Siemers, Eric and Stern, Yaakov and Yaffe, Kristine and Carrillo, Maria C. and Thies, Bill and Morrison-Bogorad, Marilyn and Wagster, Miriam V. and Phelps, Creighton H.},
  title   = {Toward defining the preclinical stages of Alzheimer's disease: recommendations from the National Institute on Aging--Alzheimer's Association workgroups on diagnostic guidelines for Alzheimer's disease},
  journal = {Alzheimer's \& Dementia},
  year    = {2011},
  volume  = {7},
  number  = {3},
  pages   = {280--292},
  date    = {2011-05},
  doi     = {10.1016/j.jalz.2011.03.003},
  pmid    = {21514248},
  pmcid   = {PMC3220946}
}

@online{NINDS_ADRD_Focus,
  author       = {{National Institute of Neurological Disorders and Stroke}},
  title        = {Focus on Alzheimer's Disease and Related Dementias},
  year         = {2025},
  month        = dec,
  day          = {12},
  url          = {https://www.ninds.nih.gov/current-research/focus-disorders/focus-alzheimers-disease-and-related-dementias},
  urldate      = {2026-02-24},
  organization = {National Institutes of Health},
  note         = {Accessed: 2026-01-25}
}

@article{edition2013diagnostic,
  title={Diagnostic and statistical manual of mental disorders},
  author={Edition, Fifth and others},
  journal={Am Psychiatric Assoc},
  volume={21},
  number={21},
  pages={591--643},
  year={2013}
}

@article{Grand2011DementiaCare,
  author  = {Grand, Jennifer H. and Caspar, Stephanie and Macdonald, Stuart W. S.},
  title   = {Clinical features and multidisciplinary approaches to dementia care},
  journal = {Journal of Multidisciplinary Healthcare},
  year    = {2011},
  volume  = {4},
  pages   = {125--147},
  date    = {2011-05-15},
  doi     = {10.2147/JMDH.S17773},
  pmid    = {21655340},
  pmcid   = {PMC3104685}
}

@article{Wang2025CognitiveAssessmentNLP_EHR,
  author  = {Wang, K. Y. and Mahesri, M. and Novoa-Laurentiev, J. and Bessette, L. G. and York, C. and Zakoul, H. and Lee, S. B. and Ngan, K. and Zhou, L. and Kim, D. H. and Lin, K. J.},
  title   = {Extracting Cognitive Impairment Assessment Information From Unstructured Notes in Electronic Health Records Using Natural Language Processing Tools: Validation with Clinical Assessment Data},
  journal = {Clinical Epidemiology},
  year    = {2025},
  volume  = {17},
  pages   = {353--365},
  date    = {2025-04-15},
  doi     = {10.2147/CLEP.S504259},
  pmid    = {40260424},
  pmcid   = {PMC12009745}
}

@misc{gemmateam2025gemma3technicalreport,
      title={Gemma 3 Technical Report}, 
      author={Gemma Team and Aishwarya Kamath and Johan Ferret and Shreya Pathak and Nino Vieillard and Ramona Merhej and Sarah Perrin and Tatiana Matejovicova and Alexandre Ramé and Morgane Rivière and Louis Rouillard and Thomas Mesnard and Geoffrey Cideron and Jean-bastien Grill and Sabela Ramos and Edouard Yvinec and Michelle Casbon and Etienne Pot and Ivo Penchev and Gaël Liu and Francesco Visin and Kathleen Kenealy and Lucas Beyer and Xiaohai Zhai and Anton Tsitsulin and Robert Busa-Fekete and Alex Feng and Noveen Sachdeva and Benjamin Coleman and Yi Gao and Basil Mustafa and Iain Barr and Emilio Parisotto and David Tian and Matan Eyal and Colin Cherry and Jan-Thorsten Peter and Danila Sinopalnikov and Surya Bhupatiraju and Rishabh Agarwal and Mehran Kazemi and Dan Malkin and Ravin Kumar and David Vilar and Idan Brusilovsky and Jiaming Luo and Andreas Steiner and Abe Friesen and Abhanshu Sharma and Abheesht Sharma and Adi Mayrav Gilady and Adrian Goedeckemeyer and Alaa Saade and Alex Feng and Alexander Kolesnikov and Alexei Bendebury and Alvin Abdagic and Amit Vadi and András György and André Susano Pinto and Anil Das and Ankur Bapna and Antoine Miech and Antoine Yang and Antonia Paterson and Ashish Shenoy and Ayan Chakrabarti and Bilal Piot and Bo Wu and Bobak Shahriari and Bryce Petrini and Charlie Chen and Charline Le Lan and Christopher A. Choquette-Choo and CJ Carey and Cormac Brick and Daniel Deutsch and Danielle Eisenbud and Dee Cattle and Derek Cheng and Dimitris Paparas and Divyashree Shivakumar Sreepathihalli and Doug Reid and Dustin Tran and Dustin Zelle and Eric Noland and Erwin Huizenga and Eugene Kharitonov and Frederick Liu and Gagik Amirkhanyan and Glenn Cameron and Hadi Hashemi and Hanna Klimczak-Plucińska and Harman Singh and Harsh Mehta and Harshal Tushar Lehri and Hussein Hazimeh and Ian Ballantyne and Idan Szpektor and Ivan Nardini and Jean Pouget-Abadie and Jetha Chan and Joe Stanton and John Wieting and Jonathan Lai and Jordi Orbay and Joseph Fernandez and Josh Newlan and Ju-yeong Ji and Jyotinder Singh and Kat Black and Kathy Yu and Kevin Hui and Kiran Vodrahalli and Klaus Greff and Linhai Qiu and Marcella Valentine and Marina Coelho and Marvin Ritter and Matt Hoffman and Matthew Watson and Mayank Chaturvedi and Michael Moynihan and Min Ma and Nabila Babar and Natasha Noy and Nathan Byrd and Nick Roy and Nikola Momchev and Nilay Chauhan and Noveen Sachdeva and Oskar Bunyan and Pankil Botarda and Paul Caron and Paul Kishan Rubenstein and Phil Culliton and Philipp Schmid and Pier Giuseppe Sessa and Pingmei Xu and Piotr Stanczyk and Pouya Tafti and Rakesh Shivanna and Renjie Wu and Renke Pan and Reza Rokni and Rob Willoughby and Rohith Vallu and Ryan Mullins and Sammy Jerome and Sara Smoot and Sertan Girgin and Shariq Iqbal and Shashir Reddy and Shruti Sheth and Siim Põder and Sijal Bhatnagar and Sindhu Raghuram Panyam and Sivan Eiger and Susan Zhang and Tianqi Liu and Trevor Yacovone and Tyler Liechty and Uday Kalra and Utku Evci and Vedant Misra and Vincent Roseberry and Vlad Feinberg and Vlad Kolesnikov and Woohyun Han and Woosuk Kwon and Xi Chen and Yinlam Chow and Yuvein Zhu and Zichuan Wei and Zoltan Egyed and Victor Cotruta and Minh Giang and Phoebe Kirk and Anand Rao and Kat Black and Nabila Babar and Jessica Lo and Erica Moreira and Luiz Gustavo Martins and Omar Sanseviero and Lucas Gonzalez and Zach Gleicher and Tris Warkentin and Vahab Mirrokni and Evan Senter and Eli Collins and Joelle Barral and Zoubin Ghahramani and Raia Hadsell and Yossi Matias and D. Sculley and Slav Petrov and Noah Fiedel and Noam Shazeer and Oriol Vinyals and Jeff Dean and Demis Hassabis and Koray Kavukcuoglu and Clement Farabet and Elena Buchatskaya and Jean-Baptiste Alayrac and Rohan Anil and Dmitry and Lepikhin and Sebastian Borgeaud and Olivier Bachem and Armand Joulin and Alek Andreev and Cassidy Hardin and Robert Dadashi and Léonard Hussenot},
      year={2025},
      eprint={2503.19786},
      archivePrefix={arXiv},
      primaryClass={cs.CL},
      url={https://arxiv.org/abs/2503.19786}, 
}

@article{McHugh2013ChiSquareIndependence,
  author  = {McHugh, Mary L.},
  title   = {The chi-square test of independence},
  journal = {Biochemia Medica (Zagreb)},
  year    = {2013},
  volume  = {23},
  number  = {2},
  pages   = {143--149},
  doi     = {10.11613/bm.2013.018},
  pmid    = {23894860},
  pmcid   = {PMC3900058}
}

@article{Tang2024EHRKnowledgeNetworksADPrediction,
  author  = {Tang, A. S. and Rankin, K. P. and Cerono, G. and Miramontes, S. and Mills, H. and Roger, J. and Zeng, B. and Nelson, C. and Soman, K. and Woldemariam, S. and Li, Y. and Lee, A. and Bove, R. and Glymour, M. and Aghaeepour, N. and Oskotsky, T. T. and Miller, Z. and Allen, I. E. and Sanders, S. J. and Baranzini, S. and Sirota, M.},
  title   = {Leveraging electronic health records and knowledge networks for {Alzheimer's} disease prediction and sex-specific biological insights},
  journal = {Nature Aging},
  year    = {2024},
  volume  = {4},
  number  = {3},
  pages   = {379--395},
  doi     = {10.1038/s43587-024-00573-8},
  pmid    = {38383858},
  pmcid   = {PMC10950787}
}

@article{Li2023EarlyPredictionADRD_EHR,
  author  = {Li, Q. and Yang, X. and Xu, J. and Guo, Y. and He, X. and Hu, H. and Lyu, T. and Marra, D. and Miller, A. and Smith, G. and DeKosky, S. and Boyce, R. D. and Schliep, K. and Shenkman, E. and Maraganore, D. and Wu, Y. and Bian, J.},
  title   = {Early prediction of {Alzheimer's} disease and related dementias using real-world electronic health records},
  journal = {Alzheimer's \& Dementia},
  year    = {2023},
  volume  = {19},
  number  = {8},
  pages   = {3506--3518},
  doi     = {10.1002/alz.12967},
  pmid    = {36815661},
  pmcid   = {PMC10976442}
}

@article{Albrecht2019AlgorithmDementiaSubtype,
  author  = {Albrecht, Julie S. and Hanna, Michael and Randall, Rachel L. and Kim, Daniel and Perfetto, Eleanor M.},
  title   = {An Algorithm to Characterize a Dementia Population by Disease Subtype},
  journal = {Alzheimer Disease and Associated Disorders},
  year    = {2019},
  volume  = {33},
  number  = {2},
  pages   = {118--123},
  doi     = {10.1097/WAD.0000000000000295},
  pmid    = {30681435},
  pmcid   = {PMC6941141}
}

@article{xie2025psci,
  title={Abstract wp175: predicting post-stroke cognitive impairment (PSCI) using multiple machine learning approaches},
  author={Xie, Yuzhang and Nahab, Fadi and Ge, Yi and Wu, Yuhua and Saurman, Jessica and Yang, Carl and Hu, Xiao},
  journal={Stroke},
  volume={56},
  number={Suppl\_1},
  pages={AWP175--AWP175},
  year={2025},
  publisher={Lippincott Williams \& Wilkins Hagerstown, MD}
}

@article{fleiss1971measuring,
  title={Measuring nominal scale agreement among many raters},
  author={Fleiss, Joseph L.},
  journal={Psychological Bulletin},
  volume={76},
  number={5},
  pages={378--382},
  year={1971},
  doi={10.1037/h0031619}
}

@article{sellergren2025medgemma,
  title={MedGemma Technical Report},
  author={Sellergren, Andrew and Kazemzadeh, Sahar and Jaroensri, Tiam and Kiraly, Atilla and Traverse, Madeleine and Kohlberger, Timo and Xu, Shawn and Jamil, Fayaz and Hughes, Cían and Lau, Charles and others},
  journal={arXiv preprint arXiv:2507.05201},
  year={2025}
}

\end{document}